# Lung Infection Quantification of COVID-19 in CT Images with Deep Learning

## Author list
Fei Shan, MD, PhD[1+], Yaozong Gao, PhD[2+], Jun Wang, PhD[3], Weiya Shi, MD[1], Nannan Shi, MD[1], Miaofei Han, MS[2], Zhong Xue, PhD[2], Dinggang Shen, PhD[2], Yuxin Shi, MD, PhD[1*]

## Institutions
[1] Department of Radiology, Shanghai Public Health Clinical Center, Fudan University, Shanghai 201508, China

[2] Department of Research and Development, Shanghai United Imaging Intelligence Co., Ltd., Shanghai 200232, China

[3] School of Communication & Information Engineering, Shanghai University, Shanghai 200444, China

+ Fei Shan and Yaozong Gao contributed equally as first authors of the paper.

## Address Correspondence to:
Yuxin Shi, MD, PhD
shiyuxin@shphc.org.cn
Department of Radiology,
Shanghai Public Health Clinical Center,
2901 Caolang Hwy,
Jinshan District,
Shanghai 201508, China

Abstract

**Background:** The quantification of COVID-19 infection in CT images using deep learning has not been investigated. Clinically there is no automatic tool to quantify the infection volume for COVID-19 patients.

**Purpose:** To develop a deep learning (DL)-based system for automatic segmentation and quantification of infection regions as well as the entire lung from chest CT scans.

**Materials and methods:** The DL-based segmentation employs the "VB-Net" neural network to segment COVID-19 infection regions in CT scans. The system is trained using 249 COVID-19 patients, and validated using 300 new COVID-19 patients. To accelerate the manual delineation of CT images for training, a human-in-the-loop (HITL) strategy is adopted to assist radiologists to refine automatic annotation of each case. To evaluate the performance of the DL-based system, Dice similarity coefficient, the differences of volume and percentage of infection (POI) are calculated between automatic and manual segmentation results on the validation set.

**Results:** The proposed system yielded Dice similarity coefficients of 91.6%±10.0% between automatic and manual segmentations, and a mean POI estimation error of 0.3% for the whole lung on the validation dataset. Moreover, compared with the cases of fully manual delineation that often takes 1 to 5 hours, the proposed human-in-the-loop strategy can dramatically reduce the delineation time to 4 minutes after 3 iterations of model updating.

**Conclusion:** A DL-based segmentation system was developed to automatically segment and quantify infection regions in CT scans of COVID-19 patients. Quantitative evaluation showed high accuracy for automatic infection region delineation, POI metrics.

## Introduction

The outbreak of 2019 novel coronavirus in Wuhan, China has rapidly spread to other countries since Dec 2019 [1-7]. The infectious disease caused by this virus was named as COVID-19 by the World Health Organization (WHO) on Feb 11, 2020 [8]. To date (Mar 5th 2020), there have been 80,565 confirmed cases in China and 95,333 confirmed cases all around the world [9]. Each suspected case needs to be confirmed by the real-time polymerase chain reaction (RT-PCR) assay of the sputum [10]. Although it is the gold standard for diagnosis, confirming COVID-19 patients using RT-PCR is time-consuming and has been reported to suffer from high false negative rates. On the other hand, because chest CT scans collected from COVID-19 patients frequently show bilateral patchy shadows or ground glass opacity (GGO) in the lung [11, 12], it has been used as an important complementary indicator in COVID-19 screening due to high sensitivity.

Chest CT examination has also shown its effectiveness in follow-up assessment of hospitalized COVID-19 patients [13]. Due to fast progression of the disease, subsequent CT scans every 3-5 days are recommended to evaluate the therapeutic responses. Although CT provides rich pathological information, only qualitative evaluation has been provided in the radiological reports owing to the lack of computerized tools to accurately quantify the infection regions and their longitudinal changes. Thus, subtle changes across follow-up CT scans are often ignored. Besides, contouring infection regions in the Chest CT is necessary for quantitative assessment; however, manual contouring of lung lesions is a tedious and time-consuming work, and inconsistent delineation could also lead to subsequent assessment discrepancies. Thus, a fast auto-contouring tool for COVID-19 infection is urgently needed in the onsite applications for quantitative disease assessment.

We developed a deep learning (DL)-based segmentation system for quantitative infection assessment. The system *not only* performs auto-contouring of infection regions, *but also* accurately estimates their shapes, volumes and percentage of infection (POI) in CT scans of COVID-19 patients. In order to provide delineation for hundreds of the training COVID-19 CT images, which is a tedious and time-consuming work, we proposed a human-in-the-loop (HITL) strategy to iteratively generate the training samples. This method involves radiologists to efficiently intervene DL-segmentation results and iteratively add more training samples to update the model, and thus greatly accelerates the algorithm development cycle. To the best of our knowledge, there are no literatures that have reported the utilization of HITL strategy in identifying COVID-19 infection in CT scans.

## Materials and Methods

### Datasets

The protocol of this retrospective study was approved by the Ethics of Committees of Shanghai Public Health Clinical Center. Informed consent was waived because of the respecpective nature of the study, and all the private information of patients was anonymized by the investigators after data collection. Totally 300 CT images from 300 COVID-19 patients (from Shanghai) were collected for validation. 249 CT images of 249 COVID-19 patients were collected from other centers (outside Shanghai) for training. The inclusion criteria are list as follows:

(a) Patients with a positive new coronavirus nucleic acid antibody and confirmed by the CDC;
(b) Patients who underwent thin-section CT;
(c) Age >=18;
(d) Presence of lung infection in CT images.

Patients with CT scans showing large motion artifacts or pre-existing lung cancer conditions were excluded in this study. 51 of 300 patients have been previously reported [23]. The prior article investigated the clinical, laboratory, and imaging findings of COVID-19 pneumonia in humans, whereas in this manuscript we develop a deep learning system to quantify COVID-19 infection in CT scans. The patient data were used for validation of system performance.

**Image Acquisition Parameters**

All COVID-19 patients underwent thin-section CT scan (SCENARIA 64 CT, Hitachi Medical, Japan). The median duration from illness onset to CT scan was 4 days, ranging from 1 to 14 days. The CT protocol was as follows: 120 kV; automatic tube current (180 mA-400 mA); iterative reconstruction; 64 mm detector; rotation time, 0.35 sec; slice thickness, 5 mm; collimation, 0.625 mm; pitch, 1.5; matrix, 512×512; and breath hold at full inspiration. The reconstruction kernel used is set as "lung smooth with a thickness of 1 mm and an interval of 0.8 mm". During reading, the mediastinal window (with window width 350 HU and window level 40 HU) and the lung window (with window width 1200 HU and window level-600 HU) were used.

**DL-Based Segmentation Network: VB-Net**

Due to the low contrast of the infection regions in CT images and large variation of both shape and position across different patients, delineating the infection regions from the chest CT scans is very challenging. We developed a DL-based network called VB-Net for this purpose. It is a modified 3-D convolutional neural network that combines V-Net [14] with the bottle-neck structure [15]. VB-Net consists of two paths (Figure 1). The first is a contracting path including down-sampling and convolution operations to extract global image features. The second is an expansive path including up-sampling and convolution operations to integrate fine-grained image features. Compared with V-Net [14], the speed of VB-Net is much faster because the bottle-neck structure is integrated in VB-Net, as detailed in Figure 1 [16, 17].

The bottle-neck design is a stacked 3-layer structure. The three layers use 1×1×1, 3×3×3 and 1×1×1 convolution kernels, where the first layer with 1×1×1 kernel reduces the number of channels and feeds the data for a regular 3×3×3 kernel layer processing, and then the channels of feature maps are restored by another 1×1×1 kernel layer. By reducing and combining feature map channels, not only the model size and inference time are greatly reduced, but also cross-channel features are effectively fused via convolusion, which makes VB-Net more applicable to deal with large 3D volumetric data than traditional V-Net.

**Training VB-Net with Human-In-The-Loop Strategy**

Training samples with detailed delineation of each infection region are required for the proposed VB-Net. However, it is a labor-intensive work for radiologists to annotate hundreds of COVID-19 CT scans. We, therefore, adopted the human-in-the-loop (HITL) strategy to iteratively update the DL model. Specifically, the training data were divided into several batches. First, the CT data in the smallest batch are manually contoured by radiologists. Then, the segmentation network was trained by this batch as an initial model. This initial model was applied to segment infection regions in the next batch, and radiologists manually correct the segmentation results provided by the segmentation network. These corrected segmentation results were then fed as new training data, and the model can be updated with increased training dataset. In this way, we iteratively increased the training dataset and built the final VB-Net. In the testing stage, the trained segmentation network segments the infection area on a new CT scan via a forward pass of neural network, and the HITL interaction also provides possible intervention and human-machine interaction for radiologists in clinical application. According to our experience, this HITL training strategy converged after 3~4 iterations. Figure 2 illustrates the process of the proposed HITL training strategy.

**Quantification and Assessment of COVID-19 Infection**

After segmentation, various metrics were computed to quantify the COVID-19 infection, including volumes of infection in the whole lung, and volumes of infection in each lobe and each bronchopulmonary segment. In addition, the POIs in the whole lung, each lobe and each bronchopulmonary segment were also computed, respecrively, to measure the severity of COVID-19 and the distribution of infection within the lung. The Hounsfield unit (HU) histogram within the infection region can also be visualized for evaluation of GGO and consolidation components inside the infection region.

Figure 3 shows the entire pipeline for quantitative COVID-19 assessment. A chest CT scan is first fed to the DL-based segmentation system, which generates infection areas, the whole lung, lung lobes, and all the bronchopulmonary segments, respectively. Then, the aforementioned quantitative metrics are calculated to quantify infection regions of the patient. The quantification provides the basis for measuring the severity of COVID-19 from the CT perspective and for tracking longitudinal changes during the treatment course.

**Statistical Analysis and Evaluation Metrics**

Statistical analysis was performed by R version 3.6.1 (R Project for Statistical Computing, Vienna, Austria). Because a majority of the continuous data did not follow a normal distribution, they were expressed as the median and interquartile range (IQR, 25th and 75th percentiles).

The Dice similarity coefficient (DSC) was used to evaluate the overlap ratio between an automatically segmented infection region ($S$) and the corresponding reference region ($R$) provided by radiologist(s). It is calculated as follows:

$$DSC(R, S) = \frac{2 \cdot |R \cap S|}{|R| + |S|},$$

where $|\cdot|$ is the operator to calculate the number of voxels in the given region, and $\cap$ is the intersection operator.

The Pearson correlation coefficient [18] was used to evaluate the correlation of two variables:

$$r = \frac{N \sum_i x_i y_i - \sum_i x_i \sum_i y_i}{\sqrt{N \sum_i x_i^2 - (\sum_i x_i)^2} \sqrt{N \sum_i y_i^2 - (\sum_i y_i)^2}},$$

where $N$ is the total number of observations, $x_i$ and $y_i$, $i = 1, \cdots, N$, are the observations of the two variables.

## Results

### Delineating Infection Regions

To demonstrate the effectiveness, Figure 4 shows typical cases of COVID-19 infection in three different stages: early stage, progressive stage and severe stage. Coronal images without and with overlaid segmentation are presented in parallel for comparison. In addition, 3D rendering of each case is also provided to give a more vivid understanding of COVID-19 infection within the lung. All three cases show that the contours delineated by the deep learning system match well with the visable lesion boundaries in CT images.

### Quantitative Evaluation on Segmentation and Measurement Accuracy

To quantitatively evaluate the accuracy of segmentation and measurement, infection regions on 300 CT scans of 300 COVID-19 patients were manually contoured by two radiologists (W.S. and F.S., with 12 and 19 years of experience in chest radiology, respectively) to serve as the reference standard. Each case was manually contoured by one radiologist and reviewed by the other. In case of disagreement, the final results were determined by consensus between the two radiologists. The automatically segmented infection regions are compared to the reference standard in terms of overlap ratio (measured by Dice similarity coefficient), volume, the percentage of infection (POI) in the whole lung, POI in each lung lobe, and POI in each bronchopulmonary segment.

Table 1 shows the statistics of these evaluations. The average Dice similarity coefficient is 91.6%±10.0% (median 92.2%, IQR 89.0%-94.6%, range 9.6%-98.1%). The average volume error is 10.7±16.7 $cm^3$ (median 5.9 $cm^3$, IQR 1.8-13.9 $cm^3$, range 0.0-89.6 $cm^3$). The mean POI estimation errors are 0.3% for the whole lung, 0.5% for lung lobes, and 0.8% for bronchopulmonary segments. 86.7% of lung-lobe POIs and 81.6% of bronchopulmonary-segment POIs are accurately estimated with differences equal or less than 1%.

Inter-rater variability was assessed by randomly sampling 10 CT scans of COVID-19 patients from the entire validation set. The two radiologists first independently contoured the infection regions in these CT scans. Their manual segmentation were then compared using the same metrics as mentioned above. Table 2 lists the quantitative comparison results. The average Dice similarity coefficient between the two radiologists is 96.1%±3.5% (median 97.2%, IQR 95.4%-98.3%, range 86.5%-99.0%). The average volume measurement difference is 7.4±5.2 $cm^3$ (median 6.8 $cm^3$, IQR 3.4-11.1 $cm^3$, range 0.2-16.3 $cm^3$). The mean POI estimation difference is 0.2% for whole lung, 0.3% for lung lobes, and 0.4% for bronchopulmonary segments. 91.4% of lung-lobe POIs and 85.9% of bronchopulmonary-segment POIs are consistently estimated with equal or less than 1% difference.

By comparing Table 1 and Table 2, it can be seen that the segmentation and measurement errors of the deep learning system are close to the inter-rater variability. This demonstrates the effectiveness of using deep learning to quantify the COVID-19 infection in CT images.

**Human-In-The-Loop Strategy**

Two metrics were used to evaluate the HITL strategy. First, the time of manual contouring was recorded to compare labeling time of a CT scan with the deep learning model. Second, the segmentation accuracy of deep learning models at different stages was assessed to see whether the accuracy improves with more annotated training data. Table 3 shows the labeling time and segmentation accuracy at different stages. Without any assistance of deep learning, it takes 211.3±52.6 minutes to contour COVID-19 infection regions on one CT scan. The contouring time drops dramatically to 31.1±8.1 minutes with the assistance of the first deep learning model trained with 36 annotated CT scans. It further drops to 12.0±2.9 minutes with 114 annotated data, and to 4.7±1.1 with 249 annotated data. Meanwhile, the segmentation accuracy of deep learning models was evaluated using Dice similarity coefficient on the entire 300 validation set. It improves from 85.1±11.4%, to 91.0±9.6%, and to 91.6%±10.0 with more training data added. The improved segmentation accuracy greatly reduces human intervention and thus reduces significantly the time of annotation and labeling.

## Discussion

CT imaging has become an efficient tool for screening COVID-19 patients and for assessing the severity of COVID-19. However, radiologists lack a computerized tool to accurately quantify the severity of COVID-19, *e.g.*, the percentage of infection in the whole lung. In the literature, deep learning has become a popular method in medical image analysis and has been used in analyzing diffuse lung diseases on CT [19, 20]. In this work, we explored deep learning to segment COVID-19 infection regions within lung fields on CT images. The accurate segmentation provides quantitative information that is necessary to track disease progression and analyze longitude changes of COVID-19 during the entire treatment period. We believe that this deep learning system for COVID-19 quantification will open up many new research directions of interest in this community.

The first potential application of this system is to quantify longitudinal changes in the follow-up CT scans of COVID-19 patients. Hospitalized patients with confirmed COVID-19 typically take a CT examination every 3-5 days. As currently there is no effective medicine to target COVID-19, most patients recover with different degrees of supportive medicine intervention. Given lots of such patients, it is interesting to see how disease progresses under different clinical management. Figure 5 gives a case with three follow-up CT scans. With infection region segmented, the changes of infection volume as well as consolidation and ground-glass opacities can be easily visualized using surface rendering technique.

The POI estimated by our system can be used to indicate the severity of COVID-19 from the radiology perspective. It is of great interst to find out how this POI correlates with clinical pneumonia assessment. Pneumonia severity index (PSI) is a clinical prediction rule that is often used to calculate the probability of morbidity and mortality among patients with community acquired pneumonia [21, 22]. It is calculated based on demographics, the coexistence of co-morbidity illnesses, and physical and laboratory examinations. In our study, COVID-19 patients were classified into non-severe (PSI level $\leq 2$) and severe groups (PSI level $\geq 3$). The POIs in the whole lung were calculated from their CT scans by the system. Based on 196 patients with both PSI and POI available, the Pearson correlation coefficient between these two variables gives 0.5, which means moderate correlation between these two scores. This result indicates the POI estimated from CT scans is clinically relevant with the severity of pneumonia. Ongoing research works are being carried on to study whether POI or its derived coefficients are helpful in predicting COVID-19 disease progression.

Another application of our system is to explore the quantitative lesion distribution specifically related to COVID-19. According to recent literature [23, 24], COVID-19 infection happens more frequently in lower lobes of the lung. However, so far no researches have reported quantitatively the severity of COVID-19 infection in each lung lobe and bronchopulmonary segment. With this deep learning system, the POIs of lung lobes and bronchopulmonary segments can be automatically calculated. Thus, statistics of infection distribution can be summarized in a large-scale dataset, *e.g.*, 300 CT scans in our study. Figure 6 show the boxplots of these POIs calculated from 300 CT scans of COVID-19 patients in Shanghai district. Figure 6(a) shows that the mean POIs of left and right lower lobes are higher than those of other lobes, which coincides with the findings reported in [23, 24].

Moreover, infection distribution can be analyzed further down to the bronchopulmonary segment level, as shown in Figure 6(b). To the best of our knowledge, this is the first work that reveals the COVID-19 distribution in bronchopulmonary segments in terms of a large-scale patient CT data. Our results show that the following segments are often infected by COVID-19 (listed with decreasing mean POI): right lower lobe - outer basal, right lower lobe – dorsal, right lower lobe – posterior basal, left lower lobe – outer basal, left lower lobe – dorsal, left lower lobe – posterior basal, and right upper lobe – back.

Using HITL strategy in training the segmentation network is a novel feature of our system. Existing AI-based systems for automatic quantitative assessment always requires a large amount of annotation CT data, whereas collecting the annotated data is very expensive or even difficult. Moreover, these AI systems are always trained as a black box to users, who however always want to know what has happened behind the model. Our experimental results indicate that the HITL strategy makes the manual annotation process faster with the assistance of deep learning models. Also, the HITL strategy makes the system more comprehensible. That is, with manual intervention in HITL, the radiologists are aware of how good the system performs in the training process. Besides, the HITL strategy helps radiologists accustomed to the AI system because they are involved in the training process. It integrates the professional knowledge from radiologists in an interactive way.

It is worth noting the limitations of our work in several aspects. First, the validation CT datasets were collected in one center, which may not be representative of all COVID-19 patients in other geographic areas. The generalization of the deep learning system needs to be further validated on multi-center datasets. Second, the system is developed to quantify infections only, and it may not be applicable for quantifying other pneumonia, *e.g.*, bacterial pneumonia. Finally, in our future work, we will extend the system to quantify severity of other pneumonia using transfer learning.

With this automatic DL-based segmentation, many studies on quantifying imaging metrics and correlating them with syndromes, epdemicology, and treatment responses could further reveal insights about imaging markers and findings towards improved diagnosis and treatment for COVID-19.

## References

1. Zhu N, Zhang D, Wang W, et al. A novel coronavirus from patients with pneumonia in China, 2019. *New England Journal of Medicine* 2020.
2. Tan W, Zhao X, Ma X, et al. A novel coronavirus genome identified in a cluster of pneumonia cases - Wuhan, China 2019 - 2020. *China CDC Weekly* 2020; **2**(4): 61-2.
3. Phan LT, Nguyen TV, Luong QC, et al. Importation and human-to-human transmission of a novel coronavirus in Vietnam. *New England Journal of Medicine* 2020; **382**(9): 872-4.
4. Holshue ML, DeBolt C, Lindquist S, et al. First case of 2019 novel coronavirus in the United States. *New England Journal of Medicine* 2020.
5. Wang C, Horby PW, Hayden FG, Gao GF. A novel coronavirus outbreak of global health concern. *The Lancet* 2020; **395**(10223): 470-3.
6. Spina S, Marrazzo F, Migliari M, Stucchi R, Sforza A, Fumagalli R. The response of Milan's Emergency Medical System to the COVID-19 outbreak in Italy. *The Lancet* 2020.
7. Wu F, Zhao S, Yu B, et al. A new coronavirus associated with human respiratory disease in China. *Nature* 2020: 1-5.
8. Gorbalenya AE. Severe acute respiratory syndrome-related coronavirus–The species and its

viruses, a statement of the Coronavirus Study Group. *BioRxiv* 2020.

9. WHO. Coronavirus disease 2019 (COVID-19)Situation Report – 45. 2020. https://www.who.int/docs/default-source/coronaviruse/situation-reports/20200305-sitrep-45-covid-19.pdf?sfvrsn=ed2ba78b_2 (accessed March 5rd 2020).

10. Xie X, Zhong Z, Zhao W, Zheng C, Wang F, Liu J. Chest CT for typical 2019-nCoV pneumonia: relationship to negative RT-PCR testing. *Radiology* 2020: 200343.

11. Huang C, Wang Y, Li X, et al. Clinical features of patients infected with 2019 novel coronavirus in Wuhan, China. *The Lancet* 2020; **395**(10223): 497-506.

12. Wang D, Hu B, Hu C, et al. Clinical characteristics of 138 hospitalized patients with 2019 novel coronavirus–infected pneumonia in Wuhan, China. *Jama* 2020.

13. Ng M-Y, Lee EY, Yang J, et al. Imaging profile of the COVID-19 infection: radiologic findings and literature review. *Radiology: Cardiothoracic Imaging* 2020; **2**(1): e200034.

14. Milletari F, Navab N, Ahmadi S-A. V-net: Fully convolutional neural networks for volumetric medical image segmentation.  2016 Fourth International Conference on 3D Vision (3DV); 2016: IEEE; 2016. p. 565-71.

15. He K, Zhang X, Ren S, Sun J. Deep residual learning for image recognition.  Proceedings of the IEEE conference on computer vision and pattern recognition; 2016; 2016. p. 770-8.

16. Han M, Zhang Y, Zhou Q, et al. Large-scale evaluation of V-Net for organ segmentation in image guided radiation therapy.  Medical Imaging 2019: Image-Guided Procedures, Robotic Interventions, and Modeling; 2019: International Society for Optics and Photonics; 2019. p. 109510O.

17. Mu G, Ma Y, Han M, Zhan Y, Zhou X, Gao Y. Automatic MR kidney segmentation for autosomal dominant polycystic kidney disease.  Medical Imaging 2019: Computer-Aided Diagnosis; 2019: International Society for Optics and Photonics; 2019. p. 109500X.

18. Cohen J, Cohen P, West SG, Aiken LS. Applied multiple regression/correlation analysis for the behavioral sciences. 3rd ed: Routledge; 2013.

19. Pang T, Guo S, Zhang X, Zhao L. Automatic Lung Segmentation Based on Texture and Deep Features of HRCT Images with Interstitial Lung Disease. *BioMed Research International* 2019; **2019**.

20. Park B, Park H, Lee SM, Seo JB, Kim N. Lung Segmentation on HRCT and Volumetric CT for Diffuse Interstitial Lung Disease Using Deep Convolutional Neural Networks. *Journal of Digital Imaging* 2019; **32**(6): 1019-26.

21. Fine MJ, Auble TE, Yealy DM, et al. A prediction rule to identify low-risk patients with community-acquired pneumonia. *New England journal of medicine* 1997; **336**(4): 243-50.

22. Shah BA, Ahmed W, Dhobi GN, Shah NN, Khursheed SQ, Haq I. Validity of pneumonia severity index and CURB-65 severity scoring systems in community acquired pneumonia in an Indian setting. *The Indian journal of chest diseases & allied sciences* 2010; **52**(1): 9-17.

23. Song F, Shi N, Shan F, et al. Emerging coronavirus 2019-nCoV pneumonia. *Radiology* 2020: 200274.

24. Bernheim A, Mei X, Huang M, et al. Chest CT Findings in Coronavirus Disease-19 (COVID-19): Relationship to Duration of Infection. *Radiology* 2020: 200463.

Tables

Table 1. Quantitative evaluation of the deep learning segmentation system on the validation dataset. The Dice coefficients, volume estimation error, and POI estimation error in the whole lung, lung lobes and bronchopulmonary segments were calculated to assess the automatic segmentation accuracy.

| Accuracy Metrics | Mean | Standard deviation | Median | 25% IQR | 75% IQR | Number of infected samples |
|---|---|---|---|---|---|---|
| Dice Similarity Coefficient | 91.6% | 10.0% | 92.2% | 89.0% | 94.6% | 300 |
| Volume Estimation Error (cm$^3$) | 10.7 | 16.7 | 5.9 | 1.8 | 13.9 | 300 |
| POI (The whole lung) | 0.3% | 0.4% | 0.1% | 0.0% | 0.4% | 300 |
| POI (Left upper lobe) | 0.4% | 1.0% | 0.1% | 0.0% | 0.4% | 233 |
| POI (Left lower lobe ) | 0.7% | 1.6% | 0.3% | 0.1% | 1.0% | 267 |
| POI (Right upper lobe) | 0.3% | 0.7% | 0.1% | 0.0% | 0.5% | 213 |
| POI (Right middle lobe) | 0.3% | 0.7% | 0.1% | 0.0% | 0.5% | 204 |
| POI (Right lower lobe) | 0.6% | 1.1% | 0.3% | 0.1% | 0.9% | 275 |
| POI (Left upper lobe / posterior tip) | 0.5% | 1.0% | 0.1% | 0.0% | 0.5% | 189 |
| POI (Left upper lobe / anterior) | 0.5% | 1.2% | 0.2% | 0.0% | 0.5% | 158 |
| POI (Left upper lobe / upper tongue) | 0.7% | 1.7% | 0.2% | 0.0% | 0.9% | 192 |
| POI (Left upper lobe / lower tongue) | 0.7% | 1.8% | 0.2% | 0.0% | 0.8% | 175 |
| POI (Left lower lobe / dorsal) | 0.9% | 2.1% | 0.4% | 0.1% | 1.2% | 224 |
| POI (Left lower lobe / anterior medial basal) | 0.6% | 1.4% | 0.2% | 0.0% | 0.8% | 209 |
| POI (Left lower lobe / outer basal) | 1.1% | 2.5% | 0.5% | 0.1% | 1.7% | 228 |
| POI (Left lower lobe / posterior basal) | 1.1% | 2.4% | 0.5% | 0.1% | 1.6% | 233 |
| POI (Right upper lobe / apical) | 0.4% | 1.1% | 0.1% | 0.0% | 0.5% | 142 |
| POI (Right upper lobe / back) | 0.7% | 1.7% | 0.2% | 0.0% | 0.8% | 186 |
| POI (Right upper lobe /anterior) | 0.4% | 1.1% | 0.1% | 0.0% | 0.9% | 151 |
| POI (Right middle lobe / lateral) | 0.6% | 1.5% | 0.1% | 0.0% | 0.6% | 183 |
| POI (Right middle lobe / medial) | 0.3% | 0.8% | 0.1% | 0.0% | 0.4% | 167 |
| POI (Right lower lobe / dorsal) | 0.9% | 1.9% | 0.4% | 0.1% | 1.4% | 233 |
| POI (Right lower lobe / inner basal) | 0.6% | 1.4% | 0.3% | 0.1% | 0.9% | 162 |
| POI (Right lower lobe / anterior basal) | 0.6% | 1.4% | 0.1% | 0.0% | 0.9% | 210 |
| POI (Right lower lobe / outer basal) | 0.9% | 1.8% | 0.4% | 0.1% | 1.2% | 236 |
| POI (Right lower lobe / posterior basal) | 1.0% | 2.0% | 0.5% | 0.1% | 1.6% | 249 |

Table 2. Inter-rater variability analysis between two radiologists on randomly sampled 10 CT cases. The Dice coefficients, volume estimation difference, and POI difference in whole lung, lung lobes and bronchopulmonary segments were estimated to serve as the reference for assessing the automatic segmentation accuracy.

| Inter-rater variability metrics | Mean | Standard deviation | Median | 25% IQR | 75% IQR | Number of infected samples |
|---|---|---|---|---|---|---|
| Dice Similarity Coefficient | 96.1% | 3.5% | 97.2% | 95.4% | 98.3% | 10 |
| Volume Estimation Error (cm$^3$) | 7.4 | 5.2 | 6.8 | 3.4 | 11.1 | 10 |
| POI (Whole lung) | 0.2% | 0.1% | 0.2% | 0.1% | 0.2% | 10 |
| POI (Left upper lobe) | 0.4% | 0.7% | 0.1% | 0.0% | 0.3% | 7 |
| POI (Left lower lobe ) | 0.2% | 0.2% | 0.3% | 0.0% | 0.4% | 7 |
| POI (Right upper lobe) | 0.3% | 0.5% | 0.1% | 0.1% | 0.3% | 6 |
| POI (Right middle lobe) | 0.3% | 0.5% | 0.1% | 0.0% | 0.1% | 6 |
| POI (Right lower lobe) | 0.2% | 0.2% | 0.2% | 0.0% | 0.3% | 9 |
| POI (Left upper lobe / posterior tip) | 0.9% | 1.1% | 0.2% | 0.0% | 1.2% | 5 |
| POI (Left upper lobe / anterior) | 0.9% | 0.8% | 0.4% | 0.3% | 1.2% | 3 |
| POI (Left upper lobe / upper tongue) | 0.6% | 0.9% | 0.0% | 0.0% | 0.6% | 7 |
| POI (Left upper lobe / lower tongue) | 0.2% | 0.2% | 0.1% | 0.0% | 0.3% | 4 |
| POI (Left lower lobe / dorsal) | 0.1% | 0.1% | 0.2% | 0.1% | 0.2% | 4 |
| POI (Left lower lobe / anterior medial basal) | 0.2% | 0.1% | 0.3% | 0.2% | 0.3% | 5 |
| POI (Left lower lobe / outer basal) | 0.3% | 0.4% | 0.2% | 0.0% | 0.4% | 6 |
| POI (Left lower lobe / posterior basal) | 0.6% | 0.5% | 0.4% | 0.2% | 0.7% | 6 |
| POI (Right upper lobe / apical) | 0.5% | 0.7% | 0.2% | 0.0% | 0.6% | 5 |
| POI (Right upper lobe / back) | 0.5% | 0.5% | 0.2% | 0.1% | 0.8% | 5 |
| POI (Right upper lobe /anterior) | 0.5% | 0.9% | 0.1% | 0.0% | 0.2% | 5 |
| POI (Right middle lobe / lateral) | 0.2% | 0.3% | 0.1% | 0.0% | 0.2% | 6 |
| POI (Right middle lobe / medial) | 0.3% | 0.6% | 0.1% | 0.0% | 0.1% | 5 |
| POI (Right lower lobe / dorsal) | 0.4% | 0.4% | 0.2% | 0.1% | 0.7% | 7 |
| POI (Right lower lobe / inner basal) | 0.5% | 0.3% | 0.6% | 0.3% | 0.8% | 4 |
| POI (Right lower lobe / anterior basal) | 0.2% | 0.3% | 0.1% | 0.0% | 0.1% | 8 |
| POI (Right lower lobe / outer basal) | 0.2% | 0.2% | 0.1% | 0.1% | 0.2% | 7 |
| POI (Right lower lobe / posterior basal) | 0.3% | 0.5% | 0.1% | 0.0% | 0.2% | 7 |

Table 3. Validation of the human-in-the-loop strategy. Manual time indicates the manual labeling/correction time without DL or with different DL models. Accuracy indicates the segmentation accuracy of DL models. # of Images indicates the number of training images used in training each DL model.

| Time (min) | Without DL | First iteration | Second iteration | Third iteration |
|---|---|---|---|---|
| Manual time | 211.3±52.6 | 31.1±8.1 | 12±2.9 | 4.7±1.1 |
| Accuracy (DSC) | N/A | 85.1±11.4% | 91.0±9.6% | 91.6%±10.0% |
| # of Images | N/A | 36 | 114 | 249 |

Figures

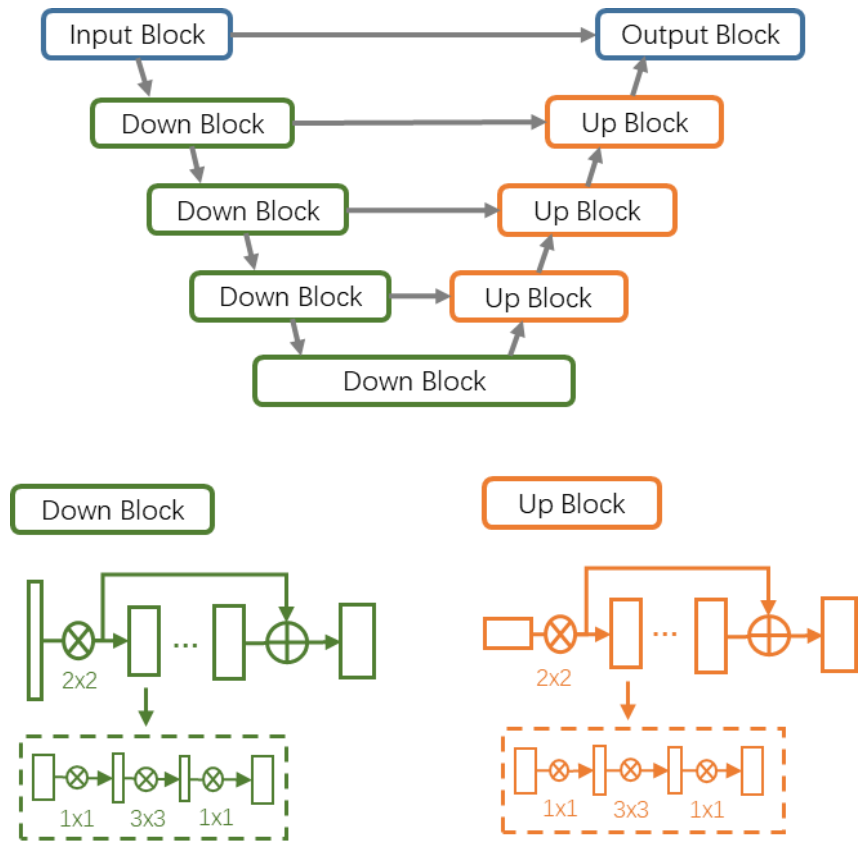

**Figure 1.** The network structure for COVID-19 infection segmentation. The dashed boxes show the bottle-neck structures inside the V-shaped network.

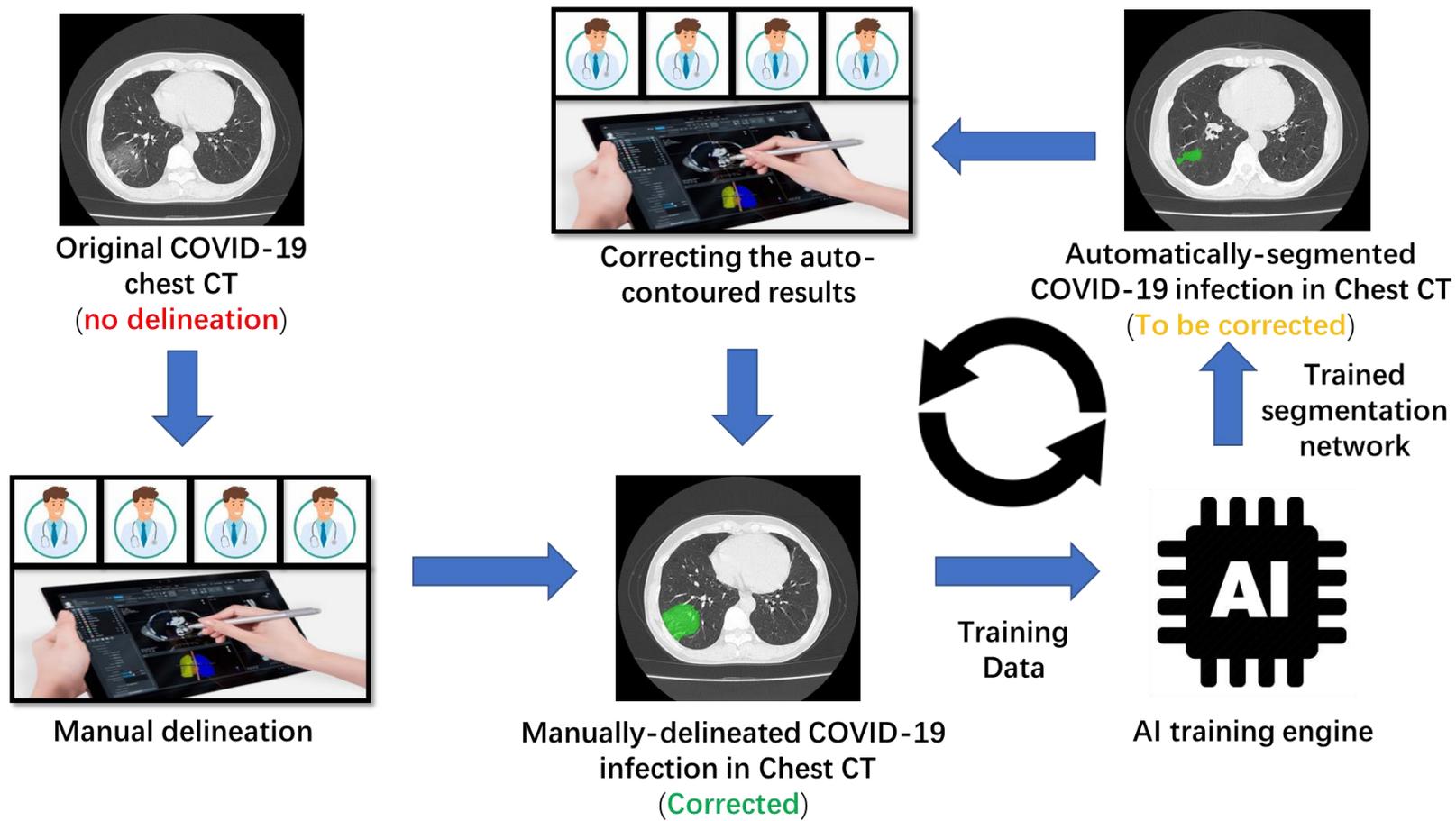

**Figure 2.** The human-in-the-loop workflow

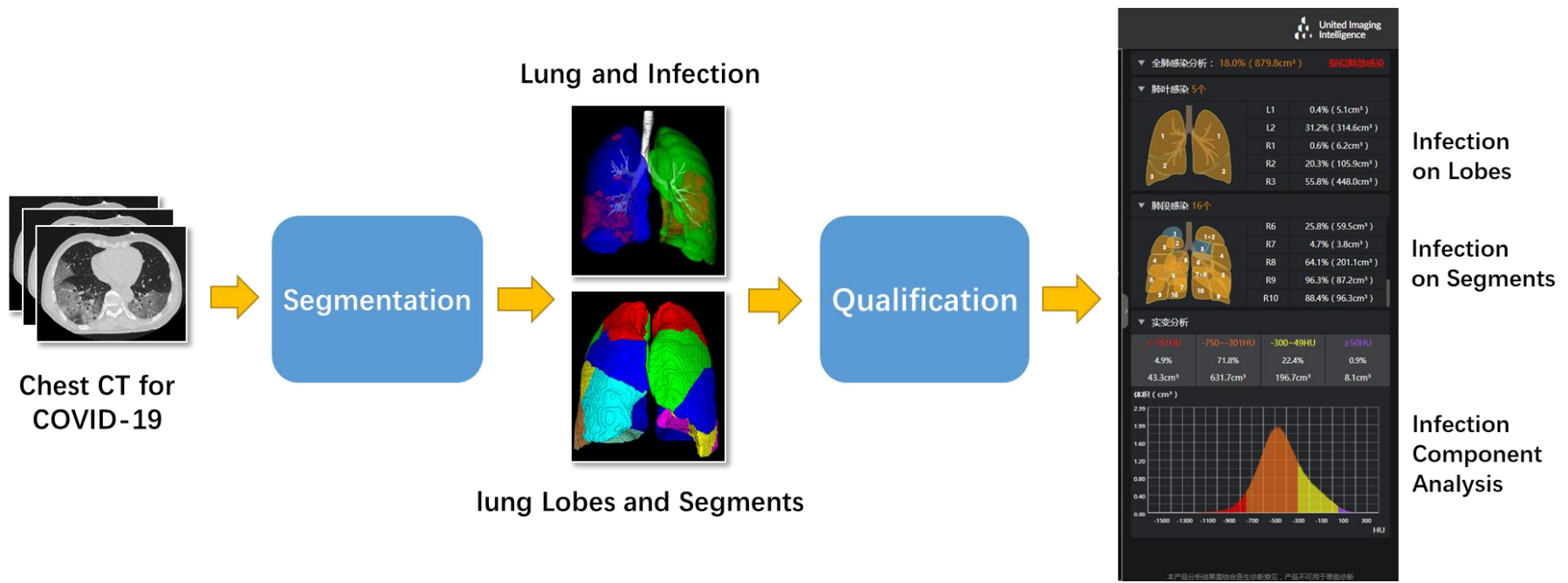

**Figure 3** Pipeline for quantifying COVID-19 infection. A chest CT scan is first fed into the DL-based segmentation system. Then, quantitative metrics are calculated to characterize infection regions in the CT scan, including but not limited to infection volumes and POIs in the whole lung, lung lobes and bronchopulmonary segments.

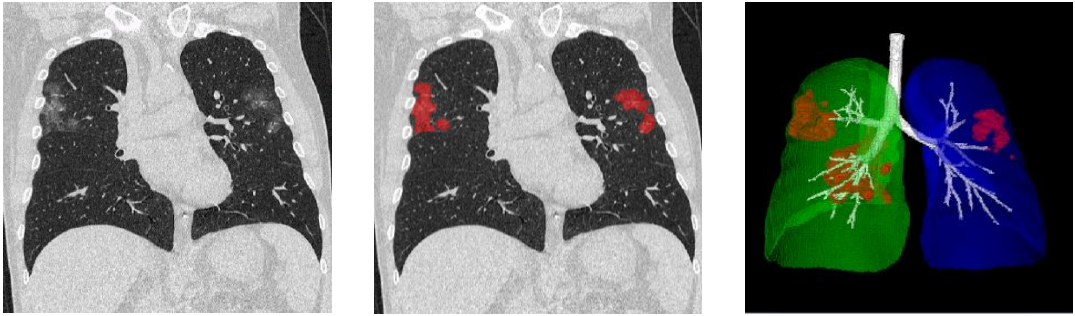
(a) A fifty-eight years old male (early stage)

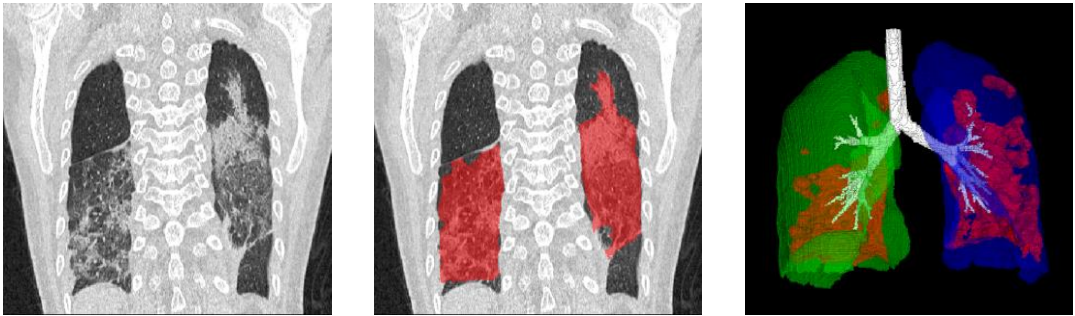
(b) A fifty-six years old female (progressive stage)

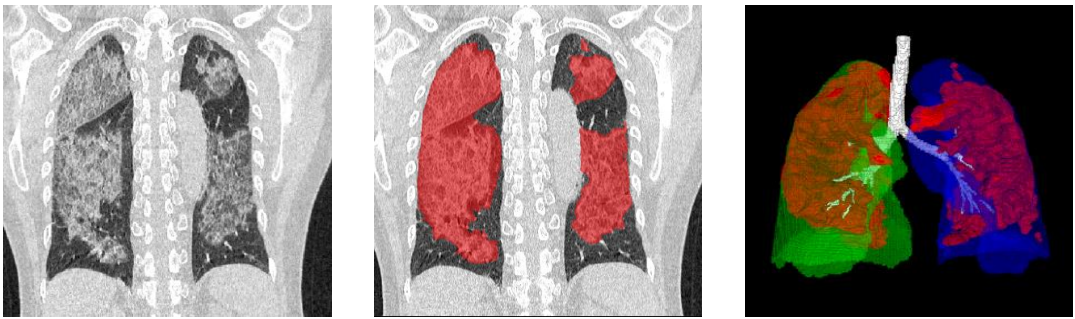
(c) A sixty-seven years old female (severe stage)

Figure 4 Typical infection segmentation results of CT scans of three COVID-19 patients. Rows 1-3: early, progressive and severe stages. Columns 1-3: CT image, CT images overlaid with segmentation, and 3D surface rendering of segmented infections.

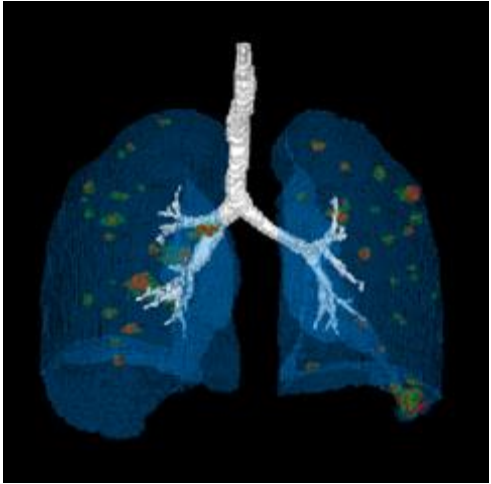
Jan 25th 2020

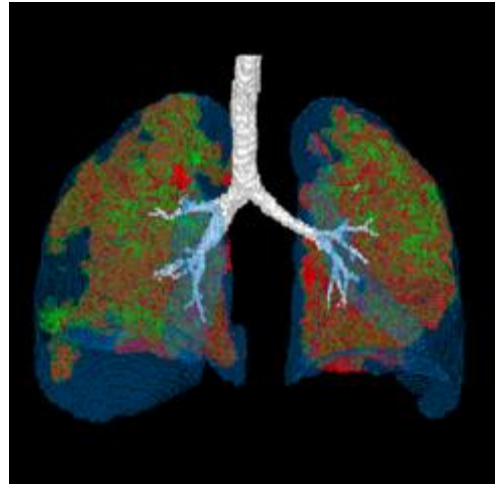
Feb 1st 2020

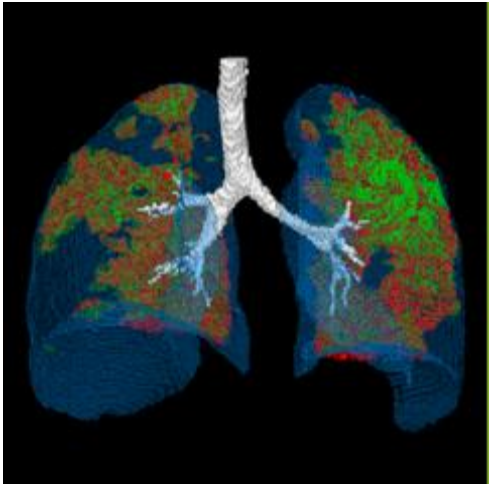
Feb 5th 2020

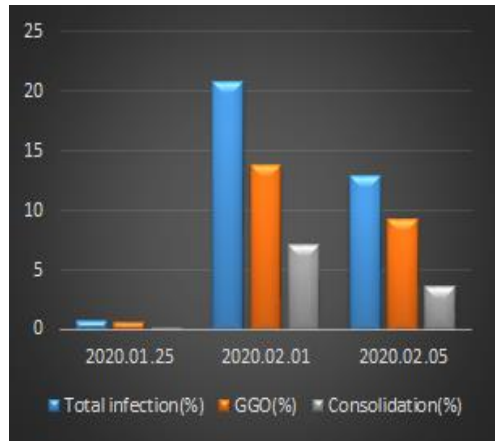
POI of GGO and consolidation regions

**Figure 5** The follow-up study results of a forty-six female patient. Green and red colors indicate ground-glass and consolidation opacities, respectively. The POI values show the progression and gradual recovery of the patient from Jan 25th, Feb 1st, to Feb 5th 2020.

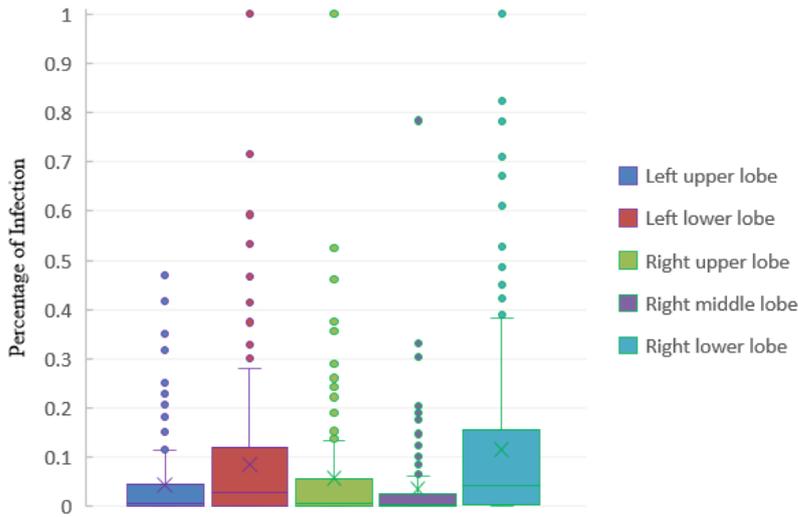

(a)

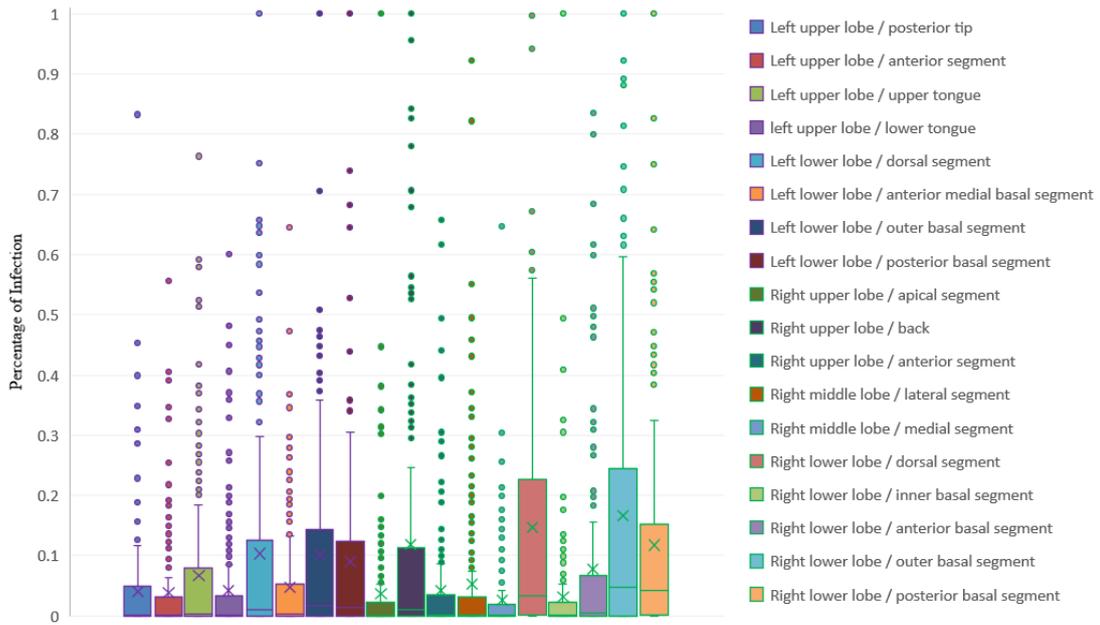

(b)

Figure 6 The box-and-whisker plots of POIs in 5 different lung lobes (a) and 18 different bronchopulmonary segments (b) on 300 validation CT scans of COVID-19 patients. The bottom and top of each box represent the 25[th] and the 75[th] percentile, respectively. The line in the box indicates the 50[th] percentile or the median value.